\documentclass[letterpaper]{article} 
\usepackage{aaai25}  
\usepackage{times}  
\usepackage{helvet}  
\usepackage{courier}  
\usepackage[hyphens]{url}  
\usepackage{amsmath}
\usepackage{graphicx} 
\usepackage{natbib}  
\usepackage{caption} 
\usepackage{algorithm}
\usepackage{algorithmic}

\usepackage{newfloat}
\usepackage{listings}
\DeclareCaptionStyle{ruled}{labelfont=normalfont,labelsep=colon,strut=off} 
\floatstyle{ruled}
\newfloat{listing}{tb}{lst}{}
\floatname{listing}{Listing}
\usepackage{kotex} 
\usepackage{todonotes}

\title{1 Trillion Token (1TT) Platform: A Novel Framework for Efficient Data Sharing and Compensation in Large Language Models}

\author{Chanjun Park\textsuperscript{\rm 1}\thanks{This author is the corresponding author.}, Hyunsoo Ha\textsuperscript{\rm 1}, Jihoo Kim\textsuperscript{\rm 1}, Yungi Kim\textsuperscript{\rm 1} \\Dahyun Kim\textsuperscript{\rm 1}, Sukyung Lee\textsuperscript{\rm 1}, Seonghoon Yang\textsuperscript{\rm 1}}

\affiliations{
\textsuperscript{\rm 1} Upstage AI \\ 
\{chanjun.park, hyunsooha, jerry, eddie, kdahyun, sukyung, hoonyang\}@upstage.ai}

\begin{document}

\maketitle

\begin{abstract}
In this paper, we propose the 1 Trillion Token Platform (1TT Platform), a novel framework designed to facilitate efficient data sharing with a transparent and equitable profit-sharing mechanism. The platform fosters collaboration between data contributors, who provide otherwise non-disclosed datasets, and a data consumer, who utilizes these datasets to enhance their own services. Data contributors are compensated in monetary terms, receiving a share of the revenue generated by the services of the data consumer. The data consumer is committed to sharing a portion of the revenue with contributors, according to predefined profit-sharing arrangements. By incorporating a transparent profit-sharing paradigm to incentivize large-scale data sharing, the 1TT Platform creates a collaborative environment to drive the advancement of NLP and LLM technologies.
\end{abstract}

%

\section{Introduction}
The availability of high-quality text data is critical for performant Natural Language Processing (NLP) services~\cite{mishra2020dqi,bhadauria2024effects}. With the rapid rise of services leveraging Large Language Models (LLMs), large-scale and high-quality text data have become ever more important~\cite{zhao2023survey,minaee2024large}. 

For acquiring such data, traditional methods include web crawling~\cite{huang2024autocrawler}, synthetic data generation~\cite{lu2023machine}, and digitizing documents via Optical Character Recognition (OCR)~\cite{islam2017survey}.
However, these methods have the following limitations: (i) web crawling may inadvertently collect copyrighted materials, leading to legal complications~\cite{krotov2020tutorial}; (ii) synthetic data generation and digitizing documents via OCR, while efficient, are constrained by the inherent capabilities of the models~\cite{liu2023hidden} used for synthetic generation and OCR, where the most performant models are often restricted for commercial use~\cite{achiam2023gpt}.

To overcome the limitations of traditional methods for data acquisition, data sharing could be a promising solution for \textit{data consumer} seeking to enhance their NLP or LLM-based services. However, it is not a straightforward issue due to \textit{the proprietary nature} of large-scale, high-quality data. To encourage \textit{data contributors} to provide such data, transparent and fair compensation must be guaranteed~\cite{berke2024insights}. To this end, we propose the 1 Trillion Token Platform (1TT Platform), a novel framework designed to facilitate seamless data sharing while ensuring transparent and fair profit-sharing mechanisms. 

The platform facilitates collaboration between \textit{data contributors} and a \textit{data consumer}. In exchange for their contributions, data contributors are compensated with a share of the revenue generated by the services of the data consumer. The data consumer agrees to allocate a portion of their revenue directly to the contributors based on predefined profit-sharing arrangements. This platform ensures that contributors are fairly compensated in monetary terms, proportionate to the success of the services utilizing their data.
Such profit sharing paradigms give incentive to share high-quality, non-disclosed data that would otherwise be hard to acquire.
By integrating secure data sharing with a transparent profit distribution system and defining clear roles for both contributors and consumer, the 1TT Platform fosters a collaborative ecosystem for advancing NLP and LLM development.

\begin{figure}[t!]
    \centering
    \includegraphics[width=0.40\textwidth]{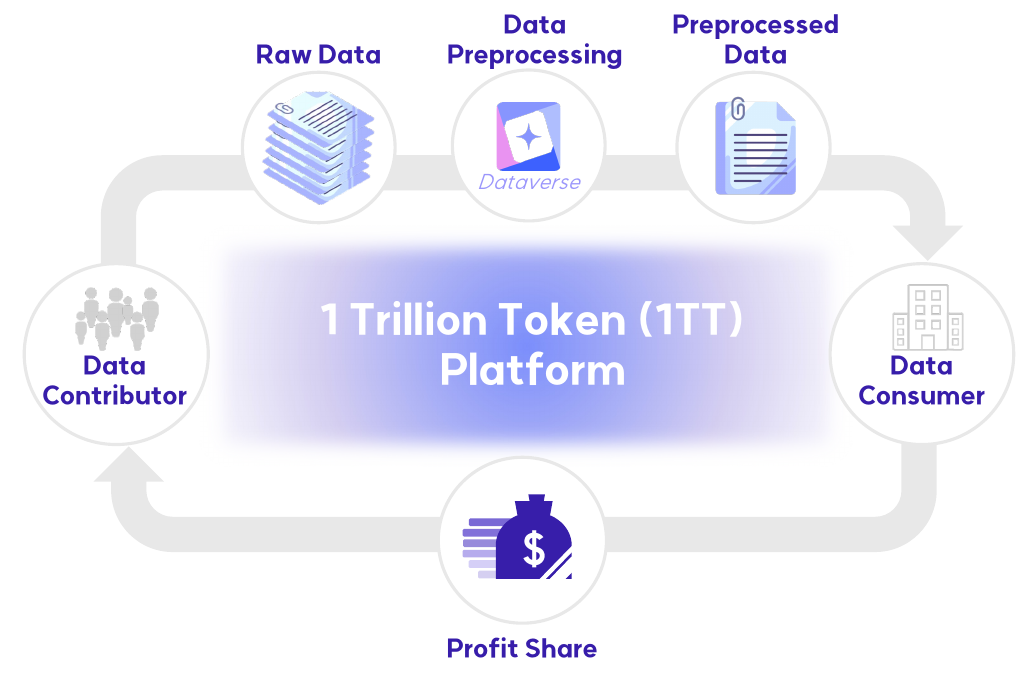}
    \caption{Overview of the 1TT Platform for data sharing and compensation. Data from the contributors are passed through data preprocessing before being used by the consumer. The data consumer shares profit with the data contributors, encouraging active data sharing.}
    \label{fig:Overview of 1T Token Platform}
\end{figure}

\begin{figure*}[t!]
    \centering
    \includegraphics[width=1.0\textwidth]{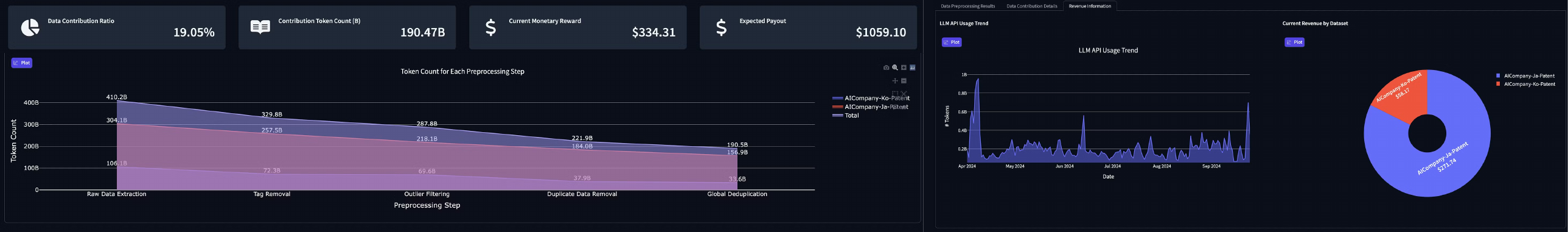}
    \caption{An application screenshot of the 1TT Platform, visible to both data contributors and consumer. The landing page illustrates key metrics such as data contribution ratio, contribution token count, current monetary reward, and expected payout.}
    \label{fig:usecase}
\end{figure*}
\section{1 Trillion Token (1TT) Platform}
\subsection{Platform Design}

As shown in Figure~\ref{fig:Overview of 1T Token Platform}, the 1TT Platform operates on a \textit{profit-sharing model} that aligns the interests of \textit{data contributors} and \textit{data consumer}. Their roles are defined as follows:
\begin{itemize} 
\item \textbf{Data Contributors}: Individuals or entities that provide datasets to the platform. 
\item \textbf{Data Consumer}: Company or organization that utilizes the contributed data to enhance their own services, generating revenue. 
\end{itemize}

When data contributors submit datasets to the platform, the contributed data undergoes an automated preprocessing phase using the open-source \texttt{Dataverse} library~\cite{park2024dataverse}, which filters these datasets based on heuristic rules and removes duplicate samples.
This preprocessing stage is essential for excluding low-quality or irrelevant data, as well as data that the consumer already possesses or that is publicly available on the web, ensuring that only meaningful contributions are compensated~\cite{zhang2023large,zhang2023jellyfish}. Then, each data contributor is compensated based on the amount of data they ultimately contribute relative to the total pool on the platform. The monetary reward $R_i$ for each contributor $i$ is calculated as follows:
\begin{equation}
    R_i = \frac{T_i}{\sum_i{T_i}} \times R_{\text{API}} \times \alpha,
\label{eq:1}
\end{equation}
where $T_i$ is the number of tokens after filtering the data contributed by contributor $i$, $R_{\text{API}}$ is the total revenue generated by the services of data consumer, and $\alpha$ is the portion of the total revenue $R_{\text{API}}$ that will be allocated to data contributors, which should be determined with consideration of the costs associated with service operations. This reward model ensures that contributors receive financial remuneration proportional to their contribution and the success of the services utilizing their data.

\subsection{Implementation}
The 1TT platform interface~\footnote{\url{https://youtu.be/aZikjbsIVY0}}, implemented using \texttt{Gradio}~\cite{abid2019gradio}, enables \textit{data contributors} to seamlessly upload datasets and provides \textit{data consumer} with the tools to manage and verify data contributions through an automated workflow.

As illustrated in Figure~\ref{fig:usecase}, key metrics, including Data Contribution Ratio, Contribution Token Count, Current Monetary Revenue, and Expected Payout, are readily available to data contributors and consumer. Expected payout is calculated based on revenue trends from the previous payout date to the present, offering interested parties an estimate of the amount they will receive on the next payment date.

In addition to these metrics, the platform offers detailed information on the following critical aspects:
\begin{itemize}
\item \textbf{Data Preprocessing Results}: Offers a transparent account of how token allocation is adjusted throughout the various stages of data preprocessing. Not only does this ensure contributors are fully informed about the handling of their submitted datasets, it also gives insights about the \textit{quality} of the submitted data for the contributors, allowing for better datasets to be shared in the future.
\item \textbf{Detailed Information about Reward}: Provides the details about the API usage patterns of data consumer and the corresponding revenue generated, allowing contributors to monitor real-time changes in their monetary rewards based on how well the services of the data consumer are faring.
\end{itemize}

By emphasizing transparency in both data processing and revenue distribution, the 1TT Platform ensures equitable compensation for contributors and fosters a sustainable ecosystem for data-driven NLP and LLM advancements.

\section{Future Work}
To enhance the 1TT Platform, future developments may include a targeted data-sharing mechanism where \textit{data consumers} submit detailed requests specifying data characteristics like domain, format, and language. This would allow \textit{data contributors} to align submissions accordingly. Additionally, a contributor reputation system could promote higher-quality contributions by helping consumers prioritize reliable sources~\cite{bouchiha2024llmchain}.

\section{Conclusion}
The 1TT Platform introduces a novel data-sharing framework with a transparent profit-sharing model for the NLP and LLM communities. By addressing key issues of fair compensation, it ensures equitable rewards for data contributors.
As a result, the platform fosters greater collaboration and long-term sustainability in advancing NLP and LLM technologies.

\section{Acknowledgments}
This work was supported by Institute of Information \& Communications Technology Planning \& Evaluation(IITP) grant funded by the Korea government(MSIT) (No. RS-2024-00338140, Development of learning and utilization technology to reflect sustainability of generative language models and up-to-dateness over time).

\bibliography{aaai25}

\end{document}